\documentclass[letterpaper, 10 pt, conference]{ieeeconf}  

\IEEEoverridecommandlockouts                              
\usepackage[style=ieee, sorting=none, citestyle=numeric-comp]{biblatex}
\usepackage[%
    colorlinks=True,
    citecolor=blue,
    pdfborder={0 0 0},
    linkcolor=red
]{hyperref}

\usepackage{algorithm} 
\usepackage{algpseudocode}

\newtheorem{definition}{Definition}

\usepackage{algorithm}
\usepackage{algorithmicx}
\usepackage{algpseudocode}
\usepackage{amsmath}

\usepackage[utf8]{inputenc}
\usepackage[T1]{fontenc}

\usepackage{graphicx}
\graphicspath{ {./Figures/} }

\usepackage{graphics} 
\usepackage{epsfig} 
\usepackage{mathptmx} 
\usepackage{times} 
\usepackage{amsmath} 

\usepackage{amssymb}  
\usepackage{bm}

\title{\LARGE \bf
Generative Adversarial Evasion and Out-of-Distribution Detection for UAV Cyber-Attacks
}

\author{Deepak Kumar Panda$^{1}$ and Weisi Guo$^{1}$
\thanks{This work was supported by the Royal Academy of Engineering and the Office of the Chief Science Adviser for National Security under the UK Intelligence Community Postdoctoral Research Fellowship programme}
\thanks{$^{1}$The authors are with Faculty of Engineering and Applied Sciences, Cranfield University, MK43 0AL Cranfield, U.K
        {\tt\small Deepak.Panda@cranfield.ac.uk, weisi.guo@cranfield.ac.uk}}%
}

\begin{document}

\maketitle
\thispagestyle{empty}
\pagestyle{empty}

\begin{abstract}
The increasing integration of UAVs into civilian airspace has amplified the urgency for resilient and intelligent intrusion detection system (IDS) frameworks, as traditional anomaly detection methods often struggle to detect novel threats. A common strategy is to treat the unfamiliar attacks as out-of-distribution (OOD) samples; hence, inadequate mitigation responses can leave systems vulnerable, granting adversaries the capability to cause potential damage. Furthermore, conventional OOD detectors frequently fail to discriminate the stealthy adversarial attacks from OOD samples. This paper proposes a conditional generative adversarial network (cGAN)-based framework specifically designed to craft stealthy adversarial attacks that effectively evade IDS mechanisms. Initially, we construct a robust multi-class classifier as IDS which classifies the benign UAV telemetry data from known cyber-attack types, including Denial of Service (DoS), false data injection (FDI), man-in-the-middle (MiTM), and replay attacks. Leveraging this classifier, our proposed cGAN strategically perturbs known attack features, generating sophisticated adversarial samples engineered to evade detection through benign misclassification. Then, the generative stealthy adversarial samples is iteratively refined to maintain statistical similarity with out-of-distribution (OOD) samples  while achieving a high attack success rate. To effectively detect these stealthy adversarial perturbations, a conditional variational autoencoder (CVAE) is implemented, using negative log-likelihood as a metric to distinguish adversarial samples from genuine OOD samples. Comparative analyses between CVAE-based regret analysis and traditional Mahalanobis distance-based detectors demonstrate that the CVAE’s negative log-likelihood significantly outperforms in detecting stealthy adversarial attacks from OOD samples. Our findings highlight the necessity of advanced probabilistic modeling techniques to reliably detect and adapt the existing IDS against novel, generative-model-based stealthy cyber threats.
\end{abstract}

\section{INTRODUCTION}
\begin{figure}[thpb]
      \centering      \includegraphics[scale=0.40]{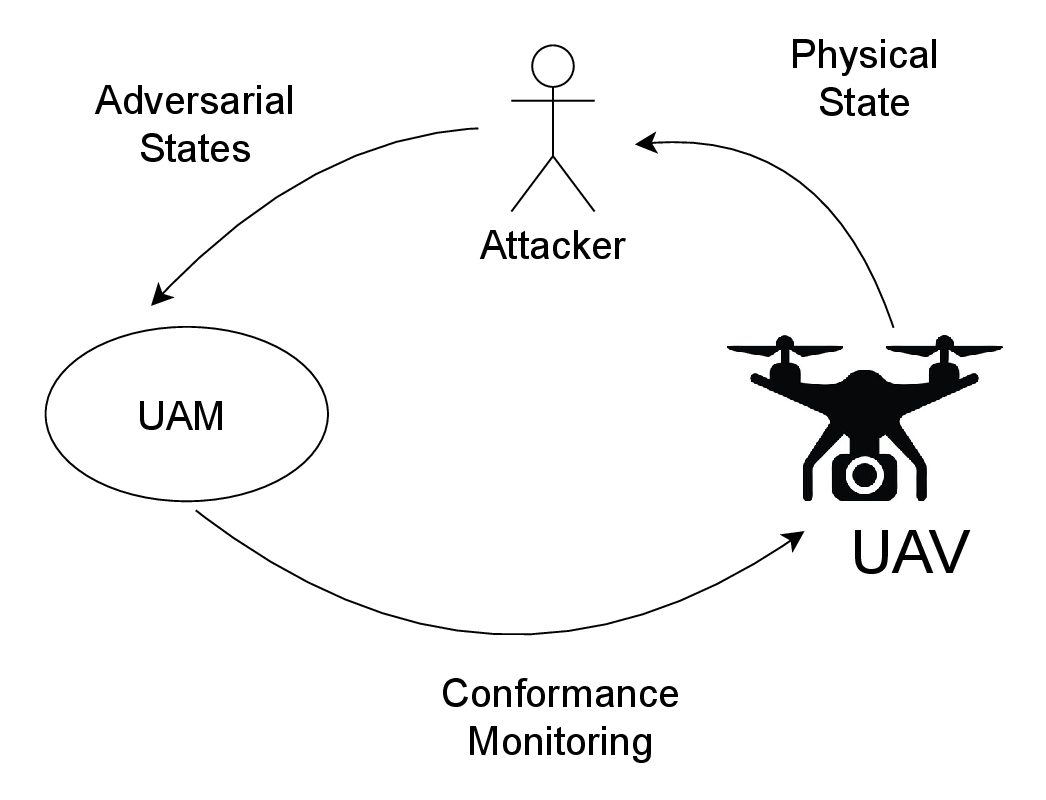}
      \caption{The attacker can manipulate the telemetry information sent by UAV to the UAM operator, which can influence the conformance monitoring for the UAV.}
    \label{figure1_app_schematic}
\end{figure}

Unmanned Aerial Vehicles (UAVs) are increasingly employed in civilian applications, including aerial surveillance, precision agriculture, and logistics \cite{shakhatreh2019unmanned}. As critical components of emerging urban air mobility (UAM) systems, UAVs rely on networked communication and onboard autonomy, rendering them vulnerable to sophisticated cyber-attacks \cite{panda2024action}. Adversaries may intercept telemetry data and inject falsified information by gaining unauthorized access to communication channels, thereby obstructing the UAM operator’s ability to verify compliance with planned flight paths as shown in Fig. \ref{figure1_app_schematic}. Failing to detect adversarial attacks as general anomalies can allow malicious actors to bypass security measures and inflict serious harm on airspace operations. This may lead to UAVs breaching restricted or unauthorized zones, thereby endangering other airspace users and heightening the risk of hijacking or unintentional exposure of sensitive information. Therefore, enhancing the security of UAV systems is crucial—especially in dynamic, time-critical operational environments. 
Traditional security methods such as cryptography and anomaly-based IDS face challenges like latency, high false positives, and poor generalization to novel threats \cite{al2024machine}.
Generative adversarial networks (GANs) have emerged as a powerful tool for crafting polymorphic cyberattacks that evade detection.  Prior work has demonstrated their use in generating adversarial examples that mimic legitimate data distributions while compromising classifier performance \cite{chauhan2020polymorphic, gu2023gan}. However, distinguishing such stealthy adversarial samples from natural out-of-distribution (OOD) samples remains a major challenge. Adversaries can exploit this weakness by generating perturbations that resemble benign anomalies, thereby bypassing anomaly detection-based IDS. An inability to differentiate out-of-distribution (OOD) samples from adversarial attacks can undermine the effectiveness of conventional filtering methods, such as the extended Kalman filter (EKF) \cite{xiao2022cyber}, which may struggle to detect stealthy adversarial inputs due to their statistical resemblance to the assumed OOD features.

Statistical methods for adversarial detection \cite{pang2018towards, smith2018understanding} often assume fixed data distributions and lack adaptability to high-dimensional stealthy attacks. Moreover, they fail to model vulnerabilities in the latent space, limiting their effectiveness in detecting unseen or well-crafted adversarial inputs. To overcome these limitations, deep generative models such as variational autoencoders (VAEs) have been proposed \cite{takiddin2021variational}. VAEs capture probabilistic latent representations that support better anomaly detection. However, their generalization ability is limited in the presence of multiple attack types. Conditional VAEs (CVAEs) \cite{pol2019anomaly}, which incorporate attack labels into the learning process, enhance discriminative power and robustness—but their application in detecting and characterizing diverse stealthy attacks remains underexplored.
To address these gaps, this paper introduces a novel framework that combines generative adversarial evasion with probabilistic detection using CVAE for UAV cyber-physical systems. The key contributions are summarized as:
\begin{itemize}
\item \textbf{Stealthy Adversarial Attack Generation:}  We design a conditional GAN that perturbs features from known cyberattacks (DoS, FDI, MiTM and replay) to produce adversarial inputs that evade a multi-class IDS by mimicking benign behavior. We optimize the stealthy adversarial attacks iteratively to ensure statistical similarity with OOD samples while achieving high evasion success.
\item \textbf{Probabilistic Detection via CVAE:} To detect such stealthy attacks from OOD samples, we implement a  CVAE-based detection system. Using negative log-likelihood as a detection score, our method outperforms traditional Mahalanobis-based detectors  \cite{pang2018towards} and advanced regret-based CVAE \cite{xiao2020likelihood} baselines in distinguishing stealthy attacks from genuine OOD samples.
\end{itemize}
\section{Overall Methodology and Dataset Description}
\begin{figure*}[thpb]
      \centering      \includegraphics[scale=0.52]{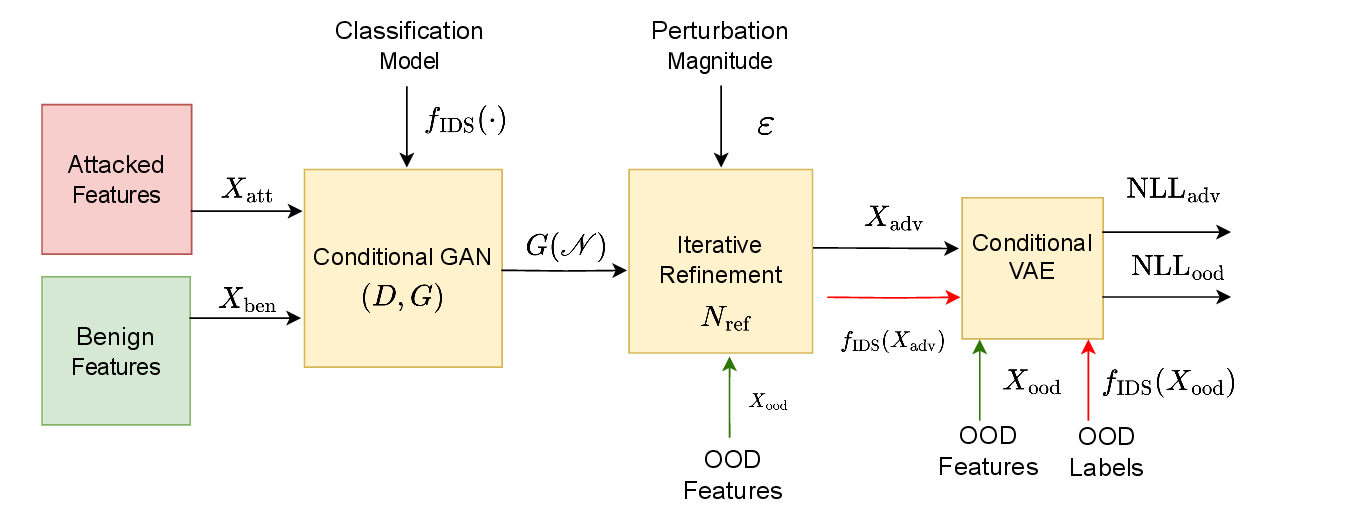}
      \caption{Schematic for obtaining negative likelihood (NLL) to detect stealthy adversarial attacks from OOD samples.}
    \label{figure1_schematic}
\end{figure*}
\subsection{Methods}
As illustrated in Fig. \ref{figure1_schematic}, the UAV cyber-physical dataset comprises benign and attacked feature sets, denoted by $X_{\textup{ben}}$ and $X_{\textup{att}}$ 
respectively. An IDS, designed as deep feedforward neural network $f_{\text{IDS}}\left( \cdot \right )$, is trained and evaluated as a multi-class classifier. Subsequently, a  cGAN is designed, consisting of a conditional generator network $G$ and a discriminator $D$. The generator $G$ learns to craft adversarial perturbations to the input attack features such that the classifier $f_{\text{IDS}} \left ( \cdot \right )$  incorrectly labels them as benign. Meanwhile, the discriminator $D$ functions as a binary classifier, distinguishing real benign samples from adversarially generated ones. This adversarial interplay compels the generator to produce increasingly stealthy and harder-to-detect perturbations. The perturbations are iteratively refined in $N_{\textup{ref}}$ steps, to produce adversarial samples $X_{\textup{adv}}$ which is statistically similar to the OOD features, but with high attack success rate. CVAE, pre-trained on both benign and attack-labeled data, is then used to compute the negative log-likelihood (NLL) for the adversarial $X_{\text{adv}}$ and OOD samples $X_{\textup{ood}}$. Based on the distribution of the negative log-likelihood the stealthy adversarial samples are classified from the OOD features. 
\subsection{Cyber Attack Dataset}
The UAV cyber-attack dataset utilized in this study is based on the intrusion detection framework developed by Hassler et al. \cite{hassler2023cyber}, where a comprehensive experimental setup was employed.  The dataset is comprised of UAV telemetry obtained after several types of cyberattacks, briefly described as follows:
\begin{itemize}
    \item \textbf{De-authentication:} Spoofed disconnects cause safety mode triggers.
    \item \textbf{Replay:} Recorded command signals are replayed to cause erratic UAV behavior.
    \item \textbf{Evil Twin:} A rogue access point enables MiTM attacks.
    \item \textbf{False Data Injection:} Sensor state manipulation misleads UAV controllers.
\end{itemize}

These attacks exhibit clear manifestations across both cyber and physical domains.  Regarding the cyber features, only those attributes common across all attack types were retained to ensure consistency. These include: timestamp, frame number, frame length, transmission duration over the wireless medium, frame sequence number, control type and subtype, and the WLAN fragment number. In contrast, the physical features encompass various UAV telemetry data, such as altitude; velocities along the $x,y,z$ axes pitch, roll, yaw; flight distance and time; pressure; battery status; and spatial distances along the $x,y,z$ axes relative to the Tello Pad. Details of the experiment is provided in \cite{hassler2023cyber}.

Given the availability of this labeled dataset, it becomes feasible to design a multi-class classification model capable of distinguishing both benign and malicious behavior and distinct attack types. However, if an adversary gains access to the labeled attack data, they can leverage this information to construct conditional GAN-based models that generate stealthy adversarial examples. These examples, synthesized from known attack features, are strategically crafted to evade existing IDS.
\section{Stealthy Adversarial Attack using GAN}
This section outlines the design of a conditional generative adversarial network (cGAN) framework aimed at generating adversarial attacks against a neural network-based (IDS) $f_{\textup{IDS}} \left ( \cdot \right )$. The cGAN  architecture comprises two neural networks: a generator $G \left (\mathfrak{z} \right )$, which synthesizes adversarial perturbations from injected noise, and a discriminator $D \left (x \right )$ which learns to differentiate between benign and adversarially perturbed inputs. These networks are trained in an adversarial (min-max) optimization framework, where the generator seeks to fool the discriminator, while the discriminator simultaneously improves its ability to distinguish between authentic and adversarial samples. The objective function is given by,
\begin{equation} \label{eq:1}
    \min_G \max_F \mathbb{E}_{x \sim {p_{\textup{data}}}} \left [ \log D \left (x \right ) \right] +  \mathbb{E}_{\mathfrak{z} \sim p_\mathfrak{z}} \left [\log \left ( 1 - D \left ( G \left ( \mathfrak{z} \right ) \right )  \right )  \right ]. 
\end{equation}
Here, $x$ represents benign data sample, $\mathfrak{z}$ represents the random noise vector, $G \left (\mathfrak{z} \right )$ generates adversarial perturbations and $D \left (x \right )$ outputs the probability that $x$ is benign. In order to solve the objective described in (\ref{eq:1}), the conditional generator and discriminator are described in the next subsection.
\subsection{Conditional Generator}
The generator is modeled as a feedforward neural network designed to transform a noise vector into adversarial feature representations. Specifically, it maps input noise to adversarial features that are misclassified as benign by the intrusion detection system (IDS). Formally, the generator is defined as a function $G : \mathbb{R}^{d_{\textup{noise}}} \rightarrow \mathbb{R}^{d_{\textup{input}}}$, where $d_{\textup{noise}}$ denotes the dimensionality of the input noise vector $\mathfrak{z}$, and $d_{\textup{input}}$ corresponds to the dimensionality of the cyber-physical feature space of the UAV system. The generator has two primary objectives: (1) to induce misclassification in the target IDS, thereby maximizing its prediction error, and (2) to ensure that the generated adversarial features closely resemble legitimate benign samples in feature space.
Let $\delta = G \left(\mathfrak{z} \right )$, represent the adversarial perturbation generated from a noise vector $\mathfrak{z}$. The adversarial input $X_{\textup{adv}}$ is then constructed by perturbing the original (attacked) feature vector with 2-norm. Let $f_{\text{IDS}} \left (X_{\text{adv}} \right ) $ denote the classification probabilities produced by the IDS when presented with adversarial input, and let $y_{\textup{ben}}$ represent the ground-truth label distribution for benign samples. 
Hence, the total loss for the cGAN generator is given by,
\begin{equation} \label{eq:2}
\begin{aligned}
\mathcal{L} &= \lambda_{\textup{cls}} \cdot \mathcal{L}_{\textup{cls}} + \lambda_{\textup{st}} \cdot \mathcal{L}_{\textup{st}} + \lambda_{G} \cdot \mathcal{L}_{G}, \\
 &= \lambda_{\textup{cls}} \cdot \frac{1}{N} \sum_{i=1}^N f_{\text{IDS}} \left (X_{\textup{adv}} \right )_i \log \frac{f_{\text{IDS}} \left (X_{\textup{adv}} \right )_i}{y_{\textup{ben}}} + \\ &\lambda_{\textup{st}} \cdot \frac{1}{N} \sum_{i=1}^{N} \left\| X_{\text{adv}} - X_{\text{ben}} \right\|_2^2 - \lambda_{G} \cdot  \mathbb{E} \left [ \log \left ( D \left (X_{\textup{adv}} \right ) \right ) \right ].
\end{aligned}
\end{equation}
where $\lambda_{\textup{cls}}$, $\lambda_{\textup{st}}$ and $\lambda_{G}$ represent hyperparameters controlling the misclassification loss, stealth loss and generator loss. The first term in (\ref{eq:2}), refers to the Kullback-Leibler (KL) divergence loss, which guides the generator towards producing benign-like predictions. The second terms, introduces a feature similarity loss, which encourages the adversarial features to remain similar to the benign features in the Eucleadian space. The third term is a part of standard cGAN formulation to help produce more benign-like samples.
\subsection{Discriminator}
The discriminator $D$ is trained to distinguish between benign and adversarial samples, hence binary cross entropy loss is used to differentiate it, which is presented as follows
\begin{equation} \label{eq:7}
    \mathcal{L}_D = -\frac{1}{2} \left( \mathbb{E}_{x \sim X_{\text{ben}}} [\log D(x)] - \mathbb{E}_{x_{\textup{adv}} \sim G} [\log (1 - D(x_{\textup{adv}}))] \right).
\end{equation}
During the cGAN training, we assume that the attacker has the access to the IDS via an external API. The cGAN can be trained in a standard manner as shown in \cite{xu2019modeling}.
\subsection{Iterative Refinement for Stealthy Attack}
Iterative refinement strategy is analogous to the one in projected gradient descent (PGD) attacks \cite{madry2018towards}, where attack is injected in multi-steps so that the overall perturbations stay within a certain bound. The aim is to make the adversarial attack $X_{\textup{adv}}$ close to the OOD sample $X_{\textup{ood}}$ generated from normally distributed noise with the noise scale $\varrho$, defined as $X_{\text{ood}} = X_{\text{att}} + \mathcal{N} \left (\mathbf{0},  {\varrho}^2 \mathbf{I} \right )$ .
 Hence, to ensure the stealth nature of the adversarial attack, $X_{\text{adv}}$ needs to be distributionally similar to $ X_{\text{ood}}$, with high attack success rate. Hence, we define the stealthy adversarial attacks as follows:
\begin{definition}  \label{stealth_attack}
Let $X \subset \mathbb{R}^d$ denote the input feature representing UAV telemetry data for IDS. Let us define $P_{\textup{ben}}$, $P_{\textup{ood}}$, $P_{\textup{adv}}$ as the distribution of the benign UAV , OOD and adversarial attacked features respectively. Let us define $\mathcal{W} \left (\cdot, \cdot \right ) $ as the Wasserstein distance measuring the distributional distance,  hence we define the stealthy adversarial attack if it satisfies the following conditions: 
\begin{equation} \label{eq:4}
\begin{gathered}
\mathcal{W} \left (P_{\textup{adv}}, P_{\textup{ben}} \right ) \approx \mathcal{W} \left (P_{\textup{ood}}, P_{\textup{ben}} \right ) , \\
\mathbb{P} \left [ f_{\textup{IDS}} \left ( X_{\textup{adv}}  \right ) \neq  f_{\textup{IDS}} \left ( X_{\textup{ben}}  \right ) \right ] \gg \mathbb{P} \left [ f_{\textup{IDS}} \left ( X_{\textup{ood}}  \right ) \neq  f_{\textup{IDS}} \left ( X_{\textup{ben}}  \right ) \right ]. 
\end{gathered}
\end{equation}
\end{definition}
If the attacker, wants to induce perturbations of maximum magnitude $\epsilon$, the definition of final adversarial attack features $X_{\text{adv}}$ under iterative refinement is given as:
\begin{equation} \label{eq:5}
\begin{gathered}
    X^{N_{\text{ref}}}_{\textup{adv}} = X_{\textup{att}} + \sum_{i=0}^{N_{\text{ref}}} \min \left ( \max \left ( X^i_{\textup{adv}} + G \left (z,c \right ) -X_{\textup{att}}, -\epsilon \right ), \epsilon  \right ), \\
    X^0_{\text{adv}} = X_{\text{att}}.
\end{gathered}
\end{equation}
If $N_{\text{att}}$ and $N^{\text{ben}}_{\text{adv}}$ represent the number of attacked and number of adversarial features classified as benign respectively, then the attack success rate is given by, $\eta_{\text{succ}} = N^{\text{ben}}_{\text{adv}} / N_{\text{att}}$. Hence, to compute the refinements $N_{\text{ref}}$ for optimum stealthy attack as per Definition \ref{stealth_attack}, the objective function can be framed as,
\begin{equation} \label{eq:6}
\begin{gathered}
\arg\min_{N_{\text{ref}}}  \mathcal{W} \left [ \mathcal{W} \left ( X_{\text{adv}}, X_{\text{att}} \right ), \mathcal{W} \left ( X_{\text{ood}}, X_{\text{att}} \right ) \right ], \\
\text{s.t.} \; \eta_{\text{succ}} \geq \eta_{\text{max}}. 
\end{gathered}
\end{equation}
The whole algorithm for generating the attack using iterative refinement is given as follows:
\begin{algorithm}[h]
\caption{Iterative Refinement of Stealthy Adversarial Samples}
\begin{algorithmic}[1]
\Require Attack input $X_{\text{att}}$, generator $G(z, c)$, bound $\epsilon$, refinement steps $N_{\text{ref}}$
\State Initialize $X^{(0)}_{\text{adv}} \gets X_{\text{att}}$
\For{$i = 1$ to $N_{\text{ref}}$}
    \State Sample noise $z \sim \mathcal{N}(0, I)$, label $c$
    \State Generate perturbation: $\delta = G(z, c)$
    \State Apply bounded update:
    \[
    X^{(i)}_{\text{adv}} \gets \text{clip}\left(X^{(i-1)}_{\text{adv}} + \delta, X_{\text{att}} - \epsilon, X_{\text{att}} + \epsilon\right)
    \]
\EndFor
\State \Return $X^{(N_{\text{ref}})}_{\text{adv}}$
\end{algorithmic}
\end{algorithm}
\section{Conditional Variational Autoencoder for Stealthy Attack Detection}
CVAEs \cite{pol2019anomaly} represent a powerful class of deep probabilistic generative models widely applied across various practical domains. These models incorporate a latent variable $\mathbf{z}$, sampled from  prior distribution $p\left ( \mathbf{z} \right )$, and  conditional distribution $p_{\theta} \left ( \mathbf{x} | \mathbf{z}, \mathbf{c} \right )$ with respect to conditional label $\mathbf{c}$  to generate the observed variable $\mathbf{x}$. However, obtaining the marginal likelihood $p_{\theta} \left ( \mathbf{x} \right )$ directly, becomes computationally intractable in high-dimensional settings due to the integration over the latent space. To address this challenge, variational inference is employed to approximate the posterior distribution and derive a tractable lower bound on the log-likelihood of the observed data. This results in the well-known evidence lower bound (ELBO), which serves as the objective function for training the model:
\begin{equation} \label{eq:7}
\begin{aligned}
\log p_\theta(\mathbf{x}) &\geq \mathbb{E}_{q_\phi(\mathbf{z}|\mathbf{x})} \left[ \log p_\theta(\mathbf{x}|\mathbf{z}, \mathbf{c}) \right] - D_{\text{KL}} \left[ q_\phi(\mathbf{z}|\mathbf{x}, \mathbf{c}) \| p(\mathbf{z}) \right] \\
&\triangleq \mathcal{L}(\mathbf{x}; \theta, \phi).
\end{aligned}
\end{equation}
Here, $q_{\phi} \left ( \mathbf{z} | \mathbf{x}, \mathbf{c} \right ) $ denotes the variational approximation to the true posterior distribution $p_{\theta} \left ( \mathbf{z} \mid \mathbf{x} \right )$, where $\phi$ and $\theta$ represent the parameters of the encoder and decoder networks, respectively. These components are typically implemented using neural networks. The CVAE is trained by minimizing the variational objective $\mathcal{L} \left ( \mathbf{x}; \theta, \phi \right ) $ over the training data. However, the CVAE does not directly maximize the true likelihood; instead, it optimizes the  ELBO, as defined in (\ref{eq:7}). This approximation can lead to learned representations that are overly constrained or limited in expressiveness, as only a limited subset of samples are likely to fall within regions of high posterior probability under the approximate posterior. To address this limitation, the importance-weighted autoencoder (IWAE) \cite{burda2015importance} is employed. The IWAE introduces a tighter lower bound on the marginal log-likelihood $\log p \left ( \mathbf{x} \right )$ by using a $k-$ sample importance-weighted estimate, thereby enabling a more expressive generative model.
\begin{equation} \label{eq:8}
    \mathcal{L}_k \left ( \mathbf{x} \right ) = \mathbb{E}_{\mathbf{z}_1, \cdots, \mathbf{z}_k \sim q\left ( \mathbf{z} \mid \mathbf{x}, \mathbf{c}  \right )}  \left [ \log \frac{1}{k} \sum_{i=1}^k \frac{p \left ( \mathbf{x}, \mathbf{z}_i\right )}{q \left ( \mathbf{z}_i | \mathbf{x}, \mathbf{c} \right )} \right ].
\end{equation}
Here $\mathbf{z}_1, \cdots, \mathbf{z}_k$ are sampled independently from the encoder network. The term inside the sum corresponds to unnormalized importance weights for joint distribution, denoted as $w_i = p\left (\mathbf{x}, \mathbf{z_i} \right ) / q \left ( \mathbf{z}_i \mid \mathbf{x} \right ) $. From Jensen's inequality, it can be shown that average importance weights are an unbiased estimator of $p \left ( \mathbf{x} \right )$ given as,
\begin{equation} \label{eq:9}
    \mathcal{L}_k = \mathbb{E} \left[ \log \frac{1}{k} \sum_{i=1}^{k} w_i \right] \leq \log \mathbb{E} \left[ \frac{1}{k} \sum_{i=1}^{k} w_i \right] = \log p(\mathbf{x}).
\end{equation}
The weight computation using importance sampling for a particular datapoint is given by, 
\begin{equation} \label{eq:10}
\begin{aligned}
    w &= \log p\left ( x | z, c \right ) + \log p \left( z \right ) - \log q  \left ( z | x,c \right ). \\
    & = \sum_{i=1}^{b} \sum_{j=1}^d x_{i,j} \log \hat{x}_{i,j} + \log p(z)  -\sum_{j=1}^{d} \left( \frac{z_j^2}{2} + \frac{\log(2\pi)}{2} \right) \\
    & -\sum_{j=1}^{d} \left( \frac{(z_j - \mu_j)^2}{2\sigma_j^2} + \frac{\log(2\pi)}{2} + \frac{\log \sigma_j^2}{2} \right).
\end{aligned}
\end{equation}
The first term represents the likelihood from the latent feature $z$ and conditional label $c$. The second term, represents the standard normally distributed prior for the latent variables $z$. The third term, represents the variational posterior for the latent variables $z$, given data $x$ and conditional label $c$. The negative log-likelihood (NLL) for the given sampled $\mathbf{x}$ is given by:
\begin{equation} \label{eq:11}
    \text{NLL} = -\left( \log \mathbb{E}_{q(z|x,c)} \left[ e^{w - \max(w)} \right] + \max(\text{w}) \right).
\end{equation}
The NLL, as shown in (\ref{eq:11}), is used to compute for the adversarial features $X_{\text{adv}}$ and OOD  features $X_{\text{ood}}$. 
\section{Results and Discussion}
\subsection{Numerical Implementation and Training Results}
\begin{figure}[thpb]
      \centering      \includegraphics[scale=0.15]{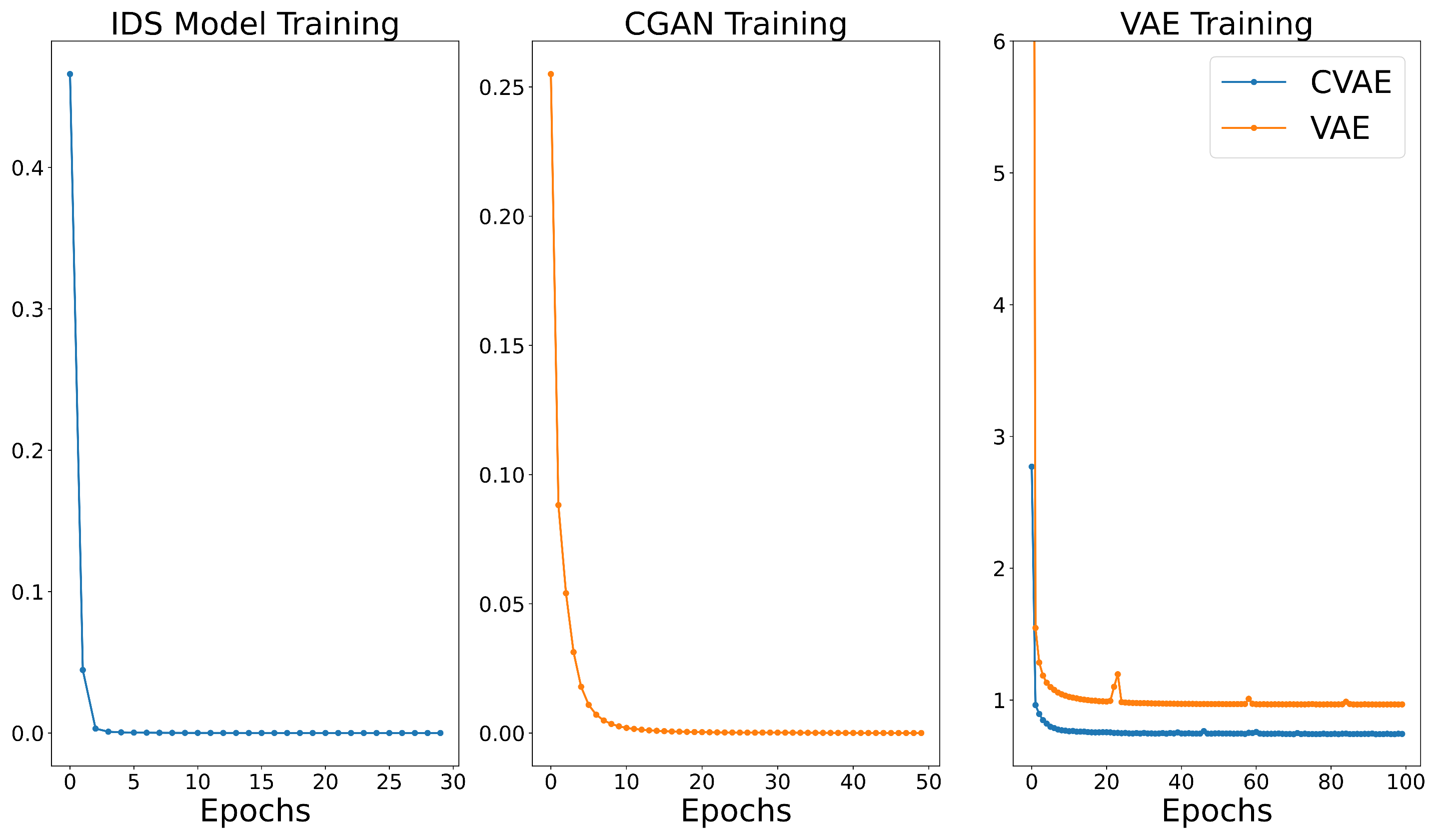}
      \caption{Loss function with the training epochs for the (a) Feedforward IDS network (b) cGAN discrimator training (c) VAE and CVAE training. }
\label{figure2_schematic}
\end{figure}
The dataset used in this study is sourced from \cite{git_UMG}, as utilized in \cite{hassler2023cyber}.  All data were normalized using the min-max scaling approach as described in \cite{hassler2023cyber}. After preprocessing, the final cyber-physical dataset comprises 12,514 instances, each with 30 features. This dataset was partitioned into training and testing subsets using an 80-20 split. To establish a baseline IDS, a multi-class feedforward neural network classifier was implemented. The network consists of three hidden layers, each containing 128 neurons. Its convergence behavior over training epochs is illustrated in Figure~\ref{figure2_schematic}a. The trained IDS achieved a perfect classification accuracy of 100\% on the test set. 

The subsequent step involved developing a cGAN to generate stealthy adversarial attack with multi-step refinement. The objective of the cGAN is to introduce adversarial perturbations into the features associated with attack samples, such that the IDS misclassifies them as benign. The cGAN architecture comprises a generator and a discriminator, each implemented as a two-hidden-layer network with 256 neurons per layer. Both components were trained using the Adam optimizer with a learning rate of 0.001. The generator loss function incorporates a weighted combination of three objectives: classification loss, stealth loss, and GAN loss. The corresponding weights were empirically set as $\left \{\lambda_{\textup{stealth}}, \lambda_{\textup{GAN}}, \lambda_{\textup{cls}}\right \} = \{ 10, 0.1, 1 \}$. The training performance of the cGAN are shown in Figure \ref{figure2_schematic}b. 

To detect the stealthy adversarial attack, both VAE and CVAE were trained to extract latent representations from the cyber-physical features. Each model was designed as a three-layer feedforward neural network with 200 latent dimensions. Performance comparison, illustrated in Figure \ref{figure2_schematic}c, indicates that the CVAE exhibits lower reconstruction loss than the standard VAE, suggesting improved capability to model the joint distribution of cyber-physical conditioned on class labels. Following the training of these networks, the final phase involves systematically characterizing adversarial attack strategies by varying the number of refinement steps and perturbation magnitudes. This analysis aims to identify regions in the input space where adversarial examples exhibit maximal stealth, thus effectively evading detection by the IDS.
\subsection{Stealthy Attack Characterization}
\begin{figure}[thpb]
      \centering      \includegraphics[scale=0.16]{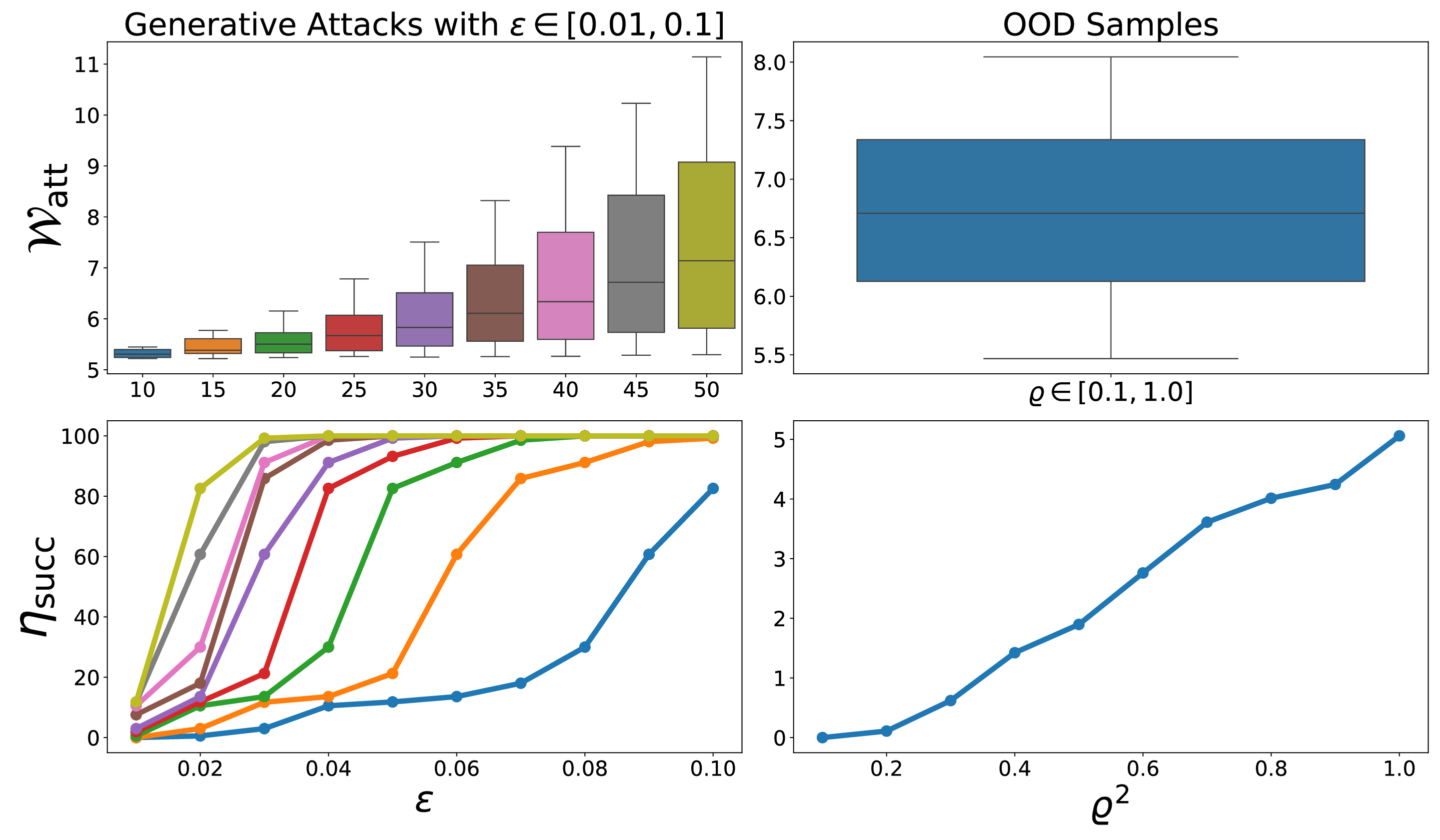}
      \caption{(a) Distributional distance between $X_{\text{adv}}$ and $X_{\text{att}}$ with various refinements. (b) Distributional distance between $X_{\text{ood}}$ and $X_{\text{att}}$ with various  refinements. (c) Attack success rate for different refinements (c) Attack success rates for different OOD samples.}
    \label{figure3_characteristic}
\end{figure}
\begin{figure}[thpb]
      \centering    \includegraphics[scale=0.10]{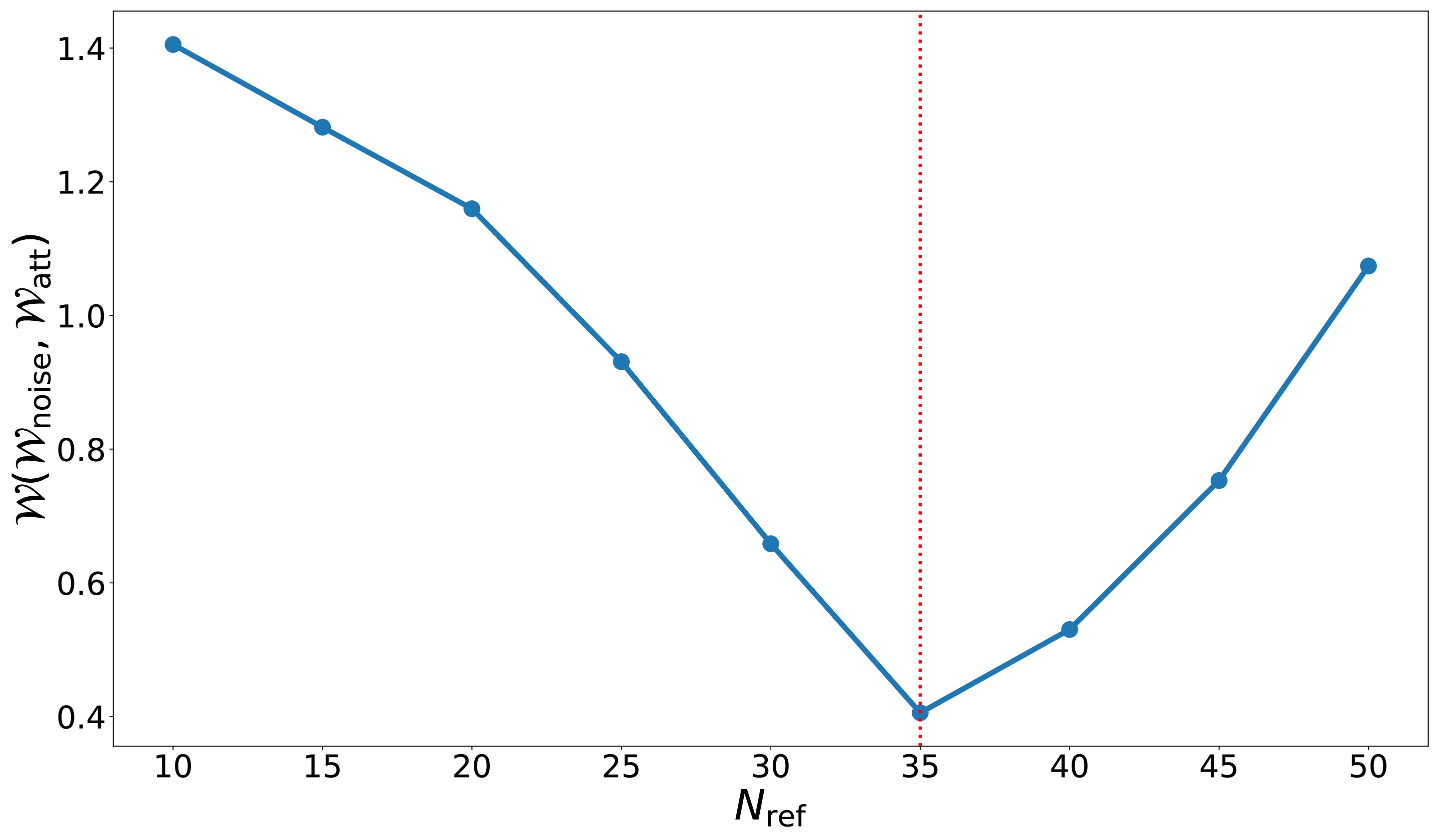}
      \caption{Distributional Distance between distributional distance between attacked and OOD samples and attacked and adversarial samples for different refinement iterations}
      \label{figure4_stealth}
\end{figure}
We assess adversarial features generated with varying refinement steps by comparing them to OOD features as per (\ref{eq:8}). The noise scale $\varrho$ is varied from 0.1 to 1, and the maximum perturbation $\epsilon$ for the iterative refinement is from 0.01 to 0.1. With too few refinement steps, features resemble OOD features and fail to consistently mislead the IDS as they can be easily detected by extended Kalman filter. Moreover,  as per Fig \ref{figure3_characteristic}, low steps causes low attack success rate, hence it does not satisfy the stealthy nature of the attack as per Definition \ref{stealth_attack}. Excessive steps for refinement, however, pushes features away from the original feature distribution, while reducing stealthy nature while introducing higher perturbation magnitude. As shown in Figure~\ref{figure3_characteristic}, attack success sharply increases beyond 30 steps, while the distributional distance between adversarial and OOD features remains minimal between 25 and 40 steps. Yet, IDS failure remains low for OOD features, unlike generative attacks, which achieve near-perfect misclassification with fewer steps. To find the optimal refinement level for stealth, we minimize the distributional distance between benign and both OOD and adversarial features, identifying 35 steps as ideal as per Fig. \ref{figure4_stealth}.If we observe Fig. \ref{figure3_characteristic}c, for $N_{\text{ref}} = 35$, the attack success rate is close to 80\% and 100\% with mere maximum perturbation magnitude of 0.03 and 0.04 respectively.
\subsection{Stealthy Attack Detection}
\begin{figure}[thpb]
      \centering
      \includegraphics[scale=0.13]{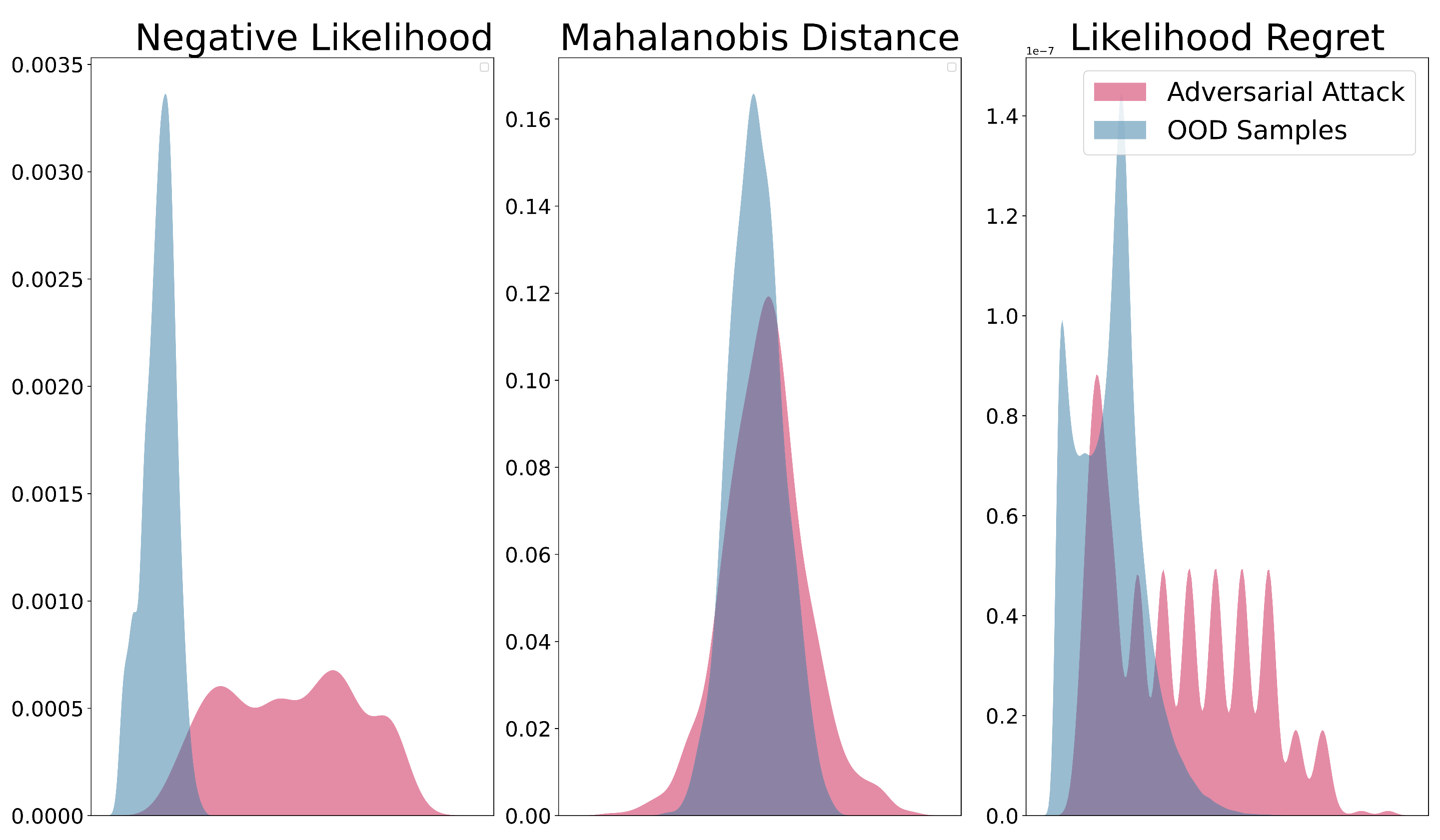}
      \caption{The metrics to detect adversarial samples from OOD samples (a) NLL (b) Mahalanobis Distance (c) Likelihood regret.}
      \label{figure4_kde}
\end{figure}
\begin{figure}[thpb]
      \centering
      \includegraphics[scale=0.14]{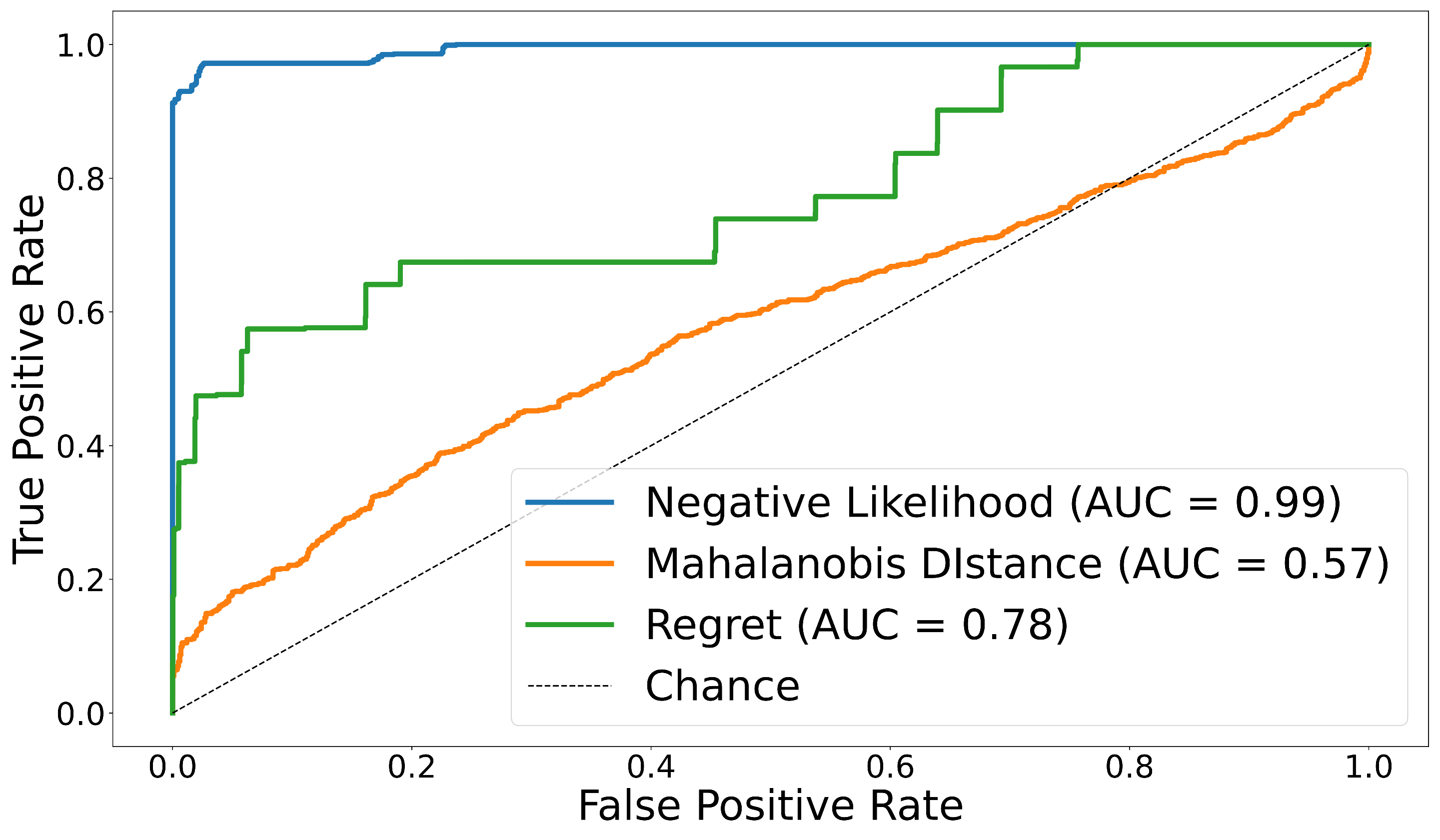}
      \caption{ROC curve for detecting adversarial samples from OOD samples.}
      \label{figure5_roc}
\end{figure}
In this subsection, we evaluate the effectiveness of different detection metrics in distinguishing stealthy generative adversarial attack from out-of-distribution (OOD) sample. As illustrated in Figure \ref{figure4_kde}, negative log-likelihood (NLL) offers superior discriminatory power between adversarial and OOD samples compared to alternative metrics such as likelihood regret \cite{xiao2020likelihood} and Mahalanobis distance \cite{pang2018towards}. While likelihood regret is capable of differentiating benign inputs from OOD samples to some extent, it fails to account for the varying characteristics of distinct OOD inputs. This limitation arises because likelihood regret relies on an relative scoring mechanism, where the reference likelihood can become unstable under various nature of OOD distribution. As a result, its sensitivity to nuanced variations in OOD distributions is reduced. Similarly, the Mahalanobis distance assumes that features within each class are drawn from well-behaved Gaussian distributions. However, this assumption does not hold in the case of CVAE, whose learned latent representations often deviate from strict Gaussianity. Empirically, this observation is supported by the area under the curve (AUC) scores, where NLL achieves a value close to 0.99 as shown in Fig. \ref{figure5_roc}.  This indicates that NLL provides a more robust and reliable signal for distinguishing both OOD and adversarial inputs, outperforming the other two metrics in this setting.  
\section{Conclusion}
In this study, we proposed a conditional GAN-based adversarial attack framework designed to generate stealthy perturbations capable of evading intrusion detection systems (IDS) within UAV environments. To improve the fidelity and evasiveness of the generated adversarial examples, we introduced an iterative  refinement mechanism that strategically minimizes the distributional distance between benign samples and both OOD and adversarial variants. Although the distributional proximity between OOD and adversarial samples relative to benign data appears similar, only the adversarial examples achieve high attack success rates—emphasizing the distinct threat posed by generative attacks. Crucially, we demonstrated that negative log-likelihood (NLL) scores derived from a conditional variational autoencoder (CVAE) offer a reliable means of detecting such stealthy adversarial inputs. In contrast, CVAE-based likelihood regret and Mahalanobis distance metrics proved insufficient for distinguishing adversarial perturbations from OOD features. These results highlight the limitations of traditional scoring approaches and underscore the importance of likelihood-based detection methods for ensuring robust and trustworthy OOD detection in secure UAV systems.


\printbibliography

\end{document}